\documentclass{llncs}

\usepackage[utf8]{inputenc}
\usepackage{amsmath}
\usepackage{graphicx}
\usepackage{amsfonts}
\usepackage{amssymb}
\usepackage{url}
\usepackage{cite}
\usepackage{multirow}
\usepackage{enumerate}
\usepackage{enumitem}
\usepackage{booktabs}
\usepackage{dirtytalk}
\usepackage{adjustbox}
\usepackage{hyperref}

\makeatletter
\newcommand{\printfnsymbol}[1]{
  \textsuperscript{\@fnsymbol{#1}}
}

\makeatother

\begin{document}

{\let\thefootnote\relax\footnotetext{Copyright \textcopyright\ 2020 for this paper by its authors. Use permitted under Creative Commons License Attribution 4.0 International (CC BY 4.0). CLEF 2020, 22-25 September 2020, Thessaloniki, Greece.}}

\title{Transferability of Natural Language Inference to Biomedical Question Answering}

\author{
Minbyul Jeong\inst{1}~\thanks{equal contribution} \and
Mujeen Sung\inst{1}\printfnsymbol{1} \and
Gangwoo Kim\inst{1} \and 
Donghyeon Kim\inst{2} \and \\
Wonjin Yoon\inst{1} \and 
Jaehyo Yoo\inst{1} \and
Jaewoo Kang\inst{1}
}
\institute{Department of Computer Science and Engineering, Korea University, Seoul, Korea \and AIR Lab, Hyundai Motor Company, Seoul, Korea \\
\email{\{minbyuljeong, mujeensung, gangwoo\_kim, wjyoon, jaehyoyoo, kangj\}@korea.ac.kr, donghyeon.kim@hyundai.com}
}

\maketitle

\begin{abstract}
Biomedical question answering (QA) is a challenging task due to the scarcity of data and the requirement of domain expertise. 
Pre-trained language models have been used to address these issues.
Recently, learning relationships between sentence pairs has been proved to improve performance in general QA.
In this paper, we focus on applying BioBERT to transfer the knowledge of natural language inference (NLI) to biomedical QA.
We observe that BioBERT trained on the NLI dataset obtains better performance on Yes/No (+5.59\%), Factoid (+0.53\%), List type (+13.58\%) questions compared to performance obtained in a previous challenge (BioASQ 7B Phase B).
We present a sequential transfer learning method that significantly performed well in the $8^{th}$ BioASQ Challenge (Phase B).
In sequential transfer learning, the order in which tasks are fine-tuned is important.
We measure an unanswerable rate of the extractive QA setting when the formats of factoid and list type questions are converted to the format of the Stanford Question Answering Dataset (SQuAD).

\end{abstract}

\section{Introduction}
Biomedical question answering (QA) is a challenging task due to the limited amount of data and the requirement of domain expertise.
To address these issues, pre-trained language models~\cite{devlin2019bert, peters2018deep} are used and further fine-tuned on a target task~\cite{hosein2019measuring, lee2019biobert, phang2018sentence, lan2019albert, talmor2019multiqa, ruder2019neural, wiese2017neural, yoon2019pre}.
Although the pre-trained language models improve performance on the target tasks, the models are still short of the upper-bound performance in biomedical QA.
Sequential transfer learning is based on transfer learning and it is used to further improve biomedical QA performance ~\cite{lee2019biobert, wiese2017neural, yoon2019pre}.
For example, fine-tuning on both the SQuAD dataset~\cite{rajpurkar2016squad} and the BioASQ dataset~\cite{tsatsaronis2015overview} results in higher performance than fine-tuning on only the BioASQ dataset.
In the general QA domain, learning relationships between sentence pairs first is effective in sequential transfer learning  ~\cite{chen2020recall, phang2018sentence, raffel2019exploring, vu2020exploring, wang2018glue}.
Thus, in this paper, we fine-tune the task of NLI~\cite{bowman2015large, williams2018broad} to improve performance in biomedical QA.
We find that performance improves when the objective function of the fine-tuned task becomes similar to the function of the downstream task.
We also find that applying the NLI task to the biomedical QA task addresses \textit{task discrepancy}.
Task discrepancy refers to the several differences in the distribution of context length, objective function, and domain shift between various fine-tuned tasks.

Specifically, we focus on reducing the discrepancy of context length distribution between NLI and biomedical QA to improve sequential transfer learning performance on the target task.
To reduce the discrepancy, we only unify the distributions of context length of the fine-tuned tasks.
We reduce the SQuAD context to a single sentence containing the ground truth answer spans~\cite{min2018efficient}.
Fine-tuning on a unified distribution reduces the time to train and perform inference on the BioASQ dataset by 52.95\% and 25\%, respectively. 
Finally, we measure an unanswerable rate of the extractive QA setting when the format of the BioASQ dataset is converted to the format of the SQuAD dataset.


Our contributions are as follows:
\begin{enumerate}[label=(\roman*)] 
\item We show that fine-tuning on an NLI dataset is effective in Yes/No, Factoid, and List type questions in BioASQ dataset.
\item We demonstrate that unifying the distributions of context length between fine-tuned tasks improves the sequential transfer learning performance of biomedical QA.
\item In the Factoid and List type questions, we measure an unanswerable rate of the extractive QA setting, when the format of the BioASQ dataset is converted to that of the SQuAD dataset.

\end{enumerate}

\section{Related Works}

\subsubsection{Transfer Learning} Transfer learning, also known as domain adaptation, refers to the situation of knowledge learned in a previous task that transferred to a subsequent task. 
In various fields including image processing or natural language processing (NLP), many studies have shown the effectiveness of transfer learning based on deep neural networks \cite{howard2018universal, long2015learning, mou2016transferable, ruder2019neural, yosinski2014transferable}. 
More recently, especially in NLP, pre-trained language models such as ELMo \cite{peters2018deep} and BERT \cite{devlin2019bert} have been used for transfer learning \cite{chen2020recall, devlin2019bert, kim2019probing, lan2019albert, liu-etal-2019-linguistic, peng2019transfer, phang2018sentence}.
In the biomedical domain, unsupervised pre-training has been used for biomedical contextualized representations \cite{beltagy2019scibert, jin2019probing, lee2019biobert}. 
BioBERT~\cite{lee2019biobert} was fine-tuned on biomedical corpora (e.g., PubMed and PubMed Central) using BERT, and BioBERT can be employed for various tasks in the biomedical or clinical domain \cite{alsentzer2019publicly, beltagy2019scibert, jin2019probing, kim2019neural, peng2019transfer, yoon2019pre}.

\subsubsection{Transferability of Natural Language Understanding}
The authors of~\cite{wiese2017neural} transferred the knowledge obtained from the SQuAD dataset to the target BioASQ dataset to address the data scarcity issue.
In \cite{lee2019biobert, yoon2019pre}, the authors adopted sequential transfer learning (e.g., \textit{BioBERT-SQuAD-BioASQ}) to improve biomedical QA performance.
Meanwhile, multiple NLI datasets have been constructed for the general domain  \cite{bowman2015large, levesque2012winograd, rajpurkar2016squad, wang2018glue, williams2018broad} and domain-specific datasets (e.g., biomedical) have recently been introduced \cite{peng2019transfer, romanov2018lessons}.
In~\cite{phang2018sentence}, the authors have found that fine-tuning on the MultiNLI (MNLI) dataset \cite{williams2018broad} consistently improves performance on target tasks in terms of all the GLUE benchmarks \cite{wang2018glue}. 
The authors of ~\cite{clark2019boolq} have found that applying knowledge from the NLI dataset improves performance on various yes and no type QA tasks in the general domain.  
Furthermore, the authors of~\cite{vu2020exploring} have used various size datasets in question answering, text classification/regression, and sequence labeling tasks.
In this paper, we use the MNLI dataset for improving performance in biomedical QA.


\section{Methods}
In this section, we outline our problem setting for the downstream task.
Our training details are provided in the Appendix~\ref{appendix:training-details}. 
We explain our method of learning biomedical entity representations using BioBERT.
Then we describe how to perform the sequential transfer learning of BioBERT for each biomedical question type of the BioASQ Challenge.
Our method can be used to apply BioBERT, which was used for training NLI dataset, to biomedical QA.

\subsection{Problem Setting}
We converted the format of the BioASQ dataset to the format of the SQuAD dataset. 
In detail, training instances in the BioASQ dataset are composed of a question (Q), human-annotated answers (A), and relevant contexts (C) (also called snippets). If the span of answers was not provided by human annotators, we first found exact spans in contexts based on human-annotated answers to factoid and list type questions. 
In this case, we enumerated all the combinations of Q-C-A triplets only when the answer span exactly matches the context. 
\textit{Yes} and \textit{No} answers to Yes/No type questions are not suggested in contexts; thus, we fine-tuned a task-specific binary classifier to predict answers.

\subsection{Overall Architecture}
Input sequence \textbf{X} consists of the concatenation of the BERT [CLS] token, Q, and C, with [SEP] tokens in between Q and C.
The sequence is denoted as \textbf{X} = \textbraceleft[CLS] $\mathbin\Vert$ Q $\mathbin\Vert$ [SEP] $\mathbin\Vert$ C $\mathbin\Vert$ [SEP]\textbraceright ~where $\mathbin\Vert$ refers to the concatenation of tensors.
The hidden representation vector of the $i^{th}$ input token is denoted as $h_i \in \mathbb{R}^{H}$ where \textit{H} denotes the hidden size.
Finally, we fine-tuned the hidden vectors corresponding to each question type, and the vectors were fed into a softmax classifier or binary classifier.

\subsubsection{Yes/No Type} 
For computing the yes probability $P^{yes}$, we projected a linear transformation matrix $M \in \mathbb{R}^{1 \times H}$ to transform the hidden representation of a [CLS] token $C \in \mathbb{R}^{H}$. 
In binary classification, the sigmoid function can be used to calculate the yes probability as follows:
\begin{equation}
    P^{yes} = \frac{1}{1+e^{-C \cdot M^\top}}
\end{equation}

The binary cross entropy loss is utilized between the yes probability $P^{yes}$ and its corresponding ground truth answer $a_{yes}$. 
Our total loss is computed as below.

\begin{equation}
    Loss = -(a_{yes}\mbox{log}P^{yes} + (1-a_{yes})\mbox{log}(1-P^{yes})) 
\end{equation}

\subsubsection{Factoid \& List type} 
At hidden representation vectors, the start and end vectors of answer spans were computed in one linear transformation matrix $M \in \mathbb{R}^{2 \times H}$. 
Let us denote the $i^{th}$ and $j^{th}$ predicted answer tokens as \textit{start} and \textit{end}, respectively. 
The probability of ($P_i^{start}, P_j^{end}$) can be calculated as follows:
\begin{equation}
    P_i = P_i^{start} \mathbin\Vert P_i^{end} = \frac{e^{h_i \cdot M^\top}}{\sum_{t=1}^{s}e^{h_t \cdot M^\top}}, 
    ~ P_j = P_j^{start} \mathbin\Vert P_j^{end} = \frac{e^{h_j \cdot M^\top}}{\sum_{t=1}^{s}e^{h_t \cdot M^\top}}
\label{eq:factoid-list-prob}
\end{equation}
where \textit{s} denotes the sequence length of BioBERT and $\cdot$ is the dot-product. 
Our objective function is the negative log-likelihood for the predicted answer with the ground truth answer position. 
Start and end position losses are computed as below:
\begin{equation}
    Loss^{start} = -\frac{1}{N}\sum_{n=1}^{N}\mbox{log}P_{a_{s}}^{start, n}, 
    ~Loss^{end} = -\frac{1}{N}\sum_{n=1}^{N}\mbox{log}P_{a_{e}}^{end, n}
\label{eq:factoid-list-loss}
\end{equation}

\noindent where \textit{N} denotes the batch size, and $a_{s}$ and $a_{e}$ are the ground truth answers of the start and end positions of each instance, respectively.
Our total loss is the arithmetic mean of $Loss^{start}$ and $Loss^{end}$.

\subsection{Transferability in domains and tasks}

\subsubsection{Yes/No Type} 
Training a model to classify relationships of sentence pairs can enhance its performance on yes or no type questions in the general domain \cite{clark2019boolq}.
Based on this finding, we believe that a classifier could be used for yes and no type questions in biomedical QA.
Thus, we fine-tuned BioBERT on the NLI task so that it can be used to answer biomedical yes or no type questions. 
We used the MNLI dataset because it is widely used and has a sufficient amount of data from various genres.
Furthermore, as shown in Table~\ref{table:nli dataset experiments 6B} and~\ref{table:nli dataset experiments 7B}, sequential transfer learning models trained on the MNLI dataset obtained meaningful results.
For our learning sequence, we fine-tuned BioBERT on the MNLI dataset which contains the relationships between hypothesis and premise sentences.
We composed a sequential transfer learning method, denoted as \textit{BioBERT-MNLI-BioASQ}.
However, using the final layer of the MNLI task instead of the binary classifier to compute $P^{yes}$ does not improve the performance of BioBERT on the BioASQ dataset.
For this reason, we added a simple binary classifier on the top layer of BioBERT.
Furthermore, the distributions of context length in the MNLI dataset and the distributions of snippets of Yes/No type questions in the BioASQ dataset are similar.
Therefore, we did not unify the context length distributions of yes and no type questions.
\vspace{-3mm}

\subsubsection{Factoid \& List Type}
The order of sequential transfer learning is important in bridging the gap between different tasks.
Performance improves when the objective function of the fine-tuned task becomes similar to that of the downstream task in Table~\ref{table:order experiment}.
Thus, we used the learning sequence \textit{BioBERT-MNLI-SQuAD-BioASQ} instead of \textit{BioBERT-SQuAD-MNLI-BioASQ}.
To address the discrepancy of context length distribution between the SQuAD dataset and the BioASQ dataset, we slightly modified the original experimental setting.
As suggested in \cite{min2018efficient}, we reorganized the context length distributions in the SQuAD dataset which is similar to the MNLI dataset and BioASQ dataset.
We developed an extractive QA setting that is scalable to minimal context and that does not use irrelevant sentences in full abstracts \cite{yoon2019pre}. 
Therefore, we extracted a sentence containing the ground truth answer span and set as a complete paragraph to construct the minimal context. 
As a result, we reduced the discrepancy of context length distribution by unifying the context length distributions for our sequential transfer learning. 
Unifying the distributions of context length reduced the time to train and perform inference on factoid and list type questions.
Our method achieved comparable results to those of the baseline method.


\section{Experiments}

\subsection{Datasets}
Our datasets are based on the pre-processed datasets provided by \cite{rajpurkar2016squad, williams2018broad, yoon2019pre}. 
For the extractive QA setting, we converted the BioASQ dataset format (Yes/No, Factoid, and List type questions) to the format of the SQuAD dataset. 
In \cite{yoon2019pre}, the authors suggested three pre-processing strategies, and for our study, we utilized two of the three strategies: \textit{Snippet-as-is} and \textit{Full-Abstract}. 
However, we added the criterion of having a blank space before and after each biomedical entity.
This criterion has shown to improve performance in distinguishing biomedical named entities.
The statistics of the pre-processed dataset are listed in Table~\ref{table:statistics}.
We have made the pre-processed BioASQ datasets publicly available.\footnote{https://github.com/dmis-lab/bioasq8b} 
In the experimental setting, we removed approximately 5K training instances from the SQuAD dataset because their answer spans do not exactly match the context.

\begin{table}[t]
\centering
\begin{adjustbox}{max width=\textwidth}
\begin{tabular}{llccc}
\toprule
\multicolumn{2}{c}{Reference System} & ~Yes/No (Macro F1)~ & ~Factoid (MRR)~ & ~List (F1)~~ \\ \midrule
\multicolumn{2}{l}{Dimitriadis \& Tsoumakas \cite{dimitriadis2019yes}} & 0.5541 & - & - \\ 
\multicolumn{2}{l}{Hosein et al., \cite{hosein2019measuring}} & - & 0.4562 & - \\ 
\multicolumn{2}{l}{Oita et al., \cite{oita2019semantically}} & 0.4831 & - & - \\
\multicolumn{2}{l}{Resta et al., \cite{resta2019transformer}} & 0.7873 & - & - \\
\multicolumn{2}{l}{Telukuntla et al., \cite{telukuntla2019uncc}} & 0.4486 & 0.4751 & 0.2002 \\
\multicolumn{2}{l}{Yoon et al., \cite{yoon2019pre}} & 0.7169 & 0.5116 & 0.4061 \\
\multicolumn{2}{l}{Ours} & \textbf{0.8432} & \textbf{0.5163} & \textbf{0.5419} \\ \bottomrule
\end{tabular}
\end{adjustbox}
\vspace{2mm}
\caption{BioASQ 7B (Phase B) Challenge results and our results. We use a dash (-) if the paper does not contain results on each question type. All the scores were averaged when the batch results are reported in each paper. In each column, the best score is in \textbf{bold}.}
\vspace{-5mm}
\label{table:7b sota}
\end{table}

\subsection{Experimental Results}
In Table~\ref{table:7b sota}, we compare our results with the best results from last year's BioASQ Challenge Task 7B (Phase B) ~\cite{dimitriadis2019yes, hosein2019measuring, oita2019semantically, resta2019transformer, telukuntla2019uncc, yoon2019pre}. 
From this comparison, we observe that training BioBERT on the MNLI dataset significantly improves its performance on the Yes/No (+5.59\%), Factoid (+0.53\%), and List (+13.58\%) type questions.

\begin{table}
\centering
\begin{adjustbox}{max width=\textwidth}
\begin{tabular}{clcccc}
\toprule
\multicolumn{6}{c}{\multirow{2}{*}{Yes/No Type}} \\
\multicolumn{6}{c}{} \\ \midrule
\multicolumn{1}{c}{\multirow{2}{*}{~\# of Tasks~}} & \multicolumn{1}{c}{\multirow{2}{*}{~Sequence of Transfer Learning~~}} & \multicolumn{4}{c}{Evaluation Metric} \\ \cmidrule{3-6} 
\multicolumn{1}{c}{} & \multicolumn{1}{c}{} & ~Accuracy~ & ~Yes F1~ & ~No F1~ & ~Macro F1~ \\ \midrule
\multirow{2}{*}{6B Test} & BioBERT-SQuAD-BioASQ & 0.8518 & 0.9004 & 0.6896 & 0.7950\\ 
 & BioBERT-MNLI-BioASQ & \textbf{0.8857} & \textbf{0.9212} & \textbf{0.7798} & \textbf{0.8505}\\ \midrule
\multirow{2}{*}{7B Test} & BioBERT-SQuAD-BioASQ &0.8595  & 0.8990 &  0.7344 & 0.8167 \\ 
 & BioBERT-MNLI-BioASQ & \textbf{0.8945} & \textbf{0.9275} & \textbf{0.7588} & \textbf{0.8432} \\ \bottomrule
\end{tabular}
\end{adjustbox}
\vspace{2mm}
\caption{Yes/No type question experiments. Evaluation metrics are accuracy (Accuracy), F1 score, and macro F1 score (Macro F1). The F1-score of yes type questions is denoted as Yes F1, and the F1 score of the no type questions is denoted as No F1. In the columns, the best score obtained in each task is in \textbf{bold}.}
\vspace{-5mm}
\label{table:yes-no type experiment}
\end{table}

First, the Yes/No type question scores obtained by our method are shown in Table~\ref{table:yes-no type experiment}.
We observed that using the SQuAD dataset for intermediate fine-tuning improves performance~\cite{lee2019biobert, wiese2017neural, yoon2019pre}.
Therefore, we evaluated our proposed method of fine-tuning BioBERT using the sequence \textit{BioBERT-SQuAD-BioASQ}, as done in~\cite{lee2019biobert, yoon2019pre}.
BioBERT is trained on the SQuAD dataset for the QA task.
Fine-tuning BioBERT with the sequence \textit{BioBERT-MNLI-BioASQ} significantly improves its performance. BioBERT obtains higher macro F1 scores (+5.55\%, +2.65\%) than the baseline. 
We believe selecting yes and no type questions in the BioASQ dataset is similar to deciding the relationship between sentence pairs in the MNLI dataset. 
We also replaced the binary classifier of BioBERT, which is trained on the BioASQ dataset, with the final layer of the MNLI task, but this did not improve performance.
Thus, we fine-tuned the binary classifier to select yes and no type questions.

\begin{table}[t]
\centering
\begin{adjustbox}{max width=\textwidth}
\begin{tabular}{clllcc|ccc}
\toprule
\multicolumn{9}{c}{\multirow{2}{*}{Context Length Discrepancy}} \\
\multicolumn{9}{c}{} \\ \midrule
\multirow{2}{*}{\# of Tasks} & \multirow{2}{*}{~~Setting} & \multicolumn{1}{c}{\multirow{2}{*}{Sequence of Transfer Learning~}} & \multicolumn{3}{c|}{Factoid (\%)} & \multicolumn{3}{c}{List (\%)} \\ \cmidrule{4-9} 
 &  & \multicolumn{1}{c}{} & \multicolumn{1}{c}{SAcc~} & LAcc~ & MRR~ & \multicolumn{1}{|c}{Prec} & Recall~ & \multicolumn{1}{c}{F1} \\ \midrule
\multirow{6}{*}{6B Test} & \multirow{2}{*}{Original} & BioBERT-SQuAD-BioASQ & 39.80 & 57.82 & 47.22 & 45.02 & \textbf{47.69} & 42.34 \\ 
 &  & BioBERT-MNLI-SQuAD-BioASQ & 38.80 & \textbf{61.34} & 47.42 & 46.60 & 47.01 & 42.44 \\ \cmidrule{2-9} 
 & \multirow{2}{*}{Document} & BioBERT-SQuAD-BioASQ & 39.71 & 56.37 & 45.81 & 46.81 & 40.26 & 39.63 \\ 
 &  & BioBERT-MNLI-SQuAD-BioASQ & 39.71 & 55.10 & 45.77 & 46.26 & 39.23 & 38.13 \\ \cmidrule{2-9} 
 & \multirow{2}{*}{Snippet} & BioBERT-SQuAD-BioASQ & 38.23 & 57.34 & 46.24 & \textbf{48.24} & 46.86 & \textbf{42.83} \\ 
 &  & BioBERT-MNLI-SQuAD-BioASQ & \textbf{41.41} & 57.40 & \textbf{48.05} & 46.01 & 45.95 & 42.75 \\ \midrule
\multirow{6}{*}{7B Test} & \multirow{2}{*}{Original} & BioBERT-SQuAD-BioASQ & 41.95 & 58.30 & 48.66 & 61.32 & 52.83 & 52.36 \\ 
 &  & BioBERT-MNLI-SQuAD-BioASQ & 42.22 & 61.06 & 49.85 & \textbf{61.46} & \textbf{54.62} & \textbf{54.19} \\ \cmidrule{2-9} 
 & \multirow{2}{*}{Document} & BioBERT-SQuAD-BioASQ & 44.46 & 57.98 & 50.02 & 58.30 & 39.19 & 43.89 \\
 &  & BioBERT-MNLI-SQuAD-BioASQ & 43.34 & 58.13 & 49.21 & 61.01 & 41.82 & 45.78 \\ \cmidrule{2-9}
 & \multirow{2}{*}{Snippet} & BioBERT-SQuAD-BioASQ & 40.79 & 58.93 & 48.27 & 60.08 & 53.96 & 53.18 \\ 
 &  & BioBERT-MNLI-SQuAD-BioASQ & \textbf{45.10} & \textbf{62.45} & \textbf{51.63} & 60.92 & 53.12 & 53.01  \\ 
 \bottomrule
\end{tabular}
\end{adjustbox}
\vspace{2mm}
\caption{Context Length Discrepancy Experiments. The metrics used to measure performance on factoid type questions are strict accuracy (SAcc), lenient accuracy (LAcc), and mean reciprocal rank (MRR). The metrics used to evaluate performance on list type questions are precision (Prec), recall (Recall), and macro F1 (F1). 'Original' indicates training BioBERT on full documents in SQuAD and snippets in BioASQ. 'Document' indicates that BioBERT was trained on full documents in SQuAD and full abstracts in BioASQ. 'Snippet' denotes training on a unified distribution of minimal context. All five batch results are averaged. In the columns, the best score obtained in each task is in \textbf{bold}.}
\vspace{-7mm}
\label{table:task discrepancy}
\end{table}

When using the factoid and list type questions in the MNLI dataset, we considered the discrepancy of context length distributions.
The obtained results are shown in Table~\ref{table:task discrepancy}. 
In the original experimental setting, full documents in the SQuAD dataset and snippets in the BioASQ dataset were used for training BioBERT.
The performance of our method on the 6B test set did not improve.
However, we observed that its performance improves with the size of the training set, as shown by the higher performance on the 7B test set compared with that on the 6B test set.

In the document setting, we used the whole paragraphs and the full abstracts of the SQuAD and BioASQ datasets, respectively.
Performance obtained in this setting is lower than that obtained in the original setting due to using longer context rather than snippets in the BioASQ dataset.
In other words, rather than using the human annotated corpus (i.e., snippets), the search space in which an answer can be found was expanded to full abstracts.
Nevertheless, the performance of our proposed method on the factoid type questions in the 7B test set improved when BioBERT was fine-tuned on the SQuAD dataset.

For the snippet setting, we unify the distributions of context length in the extractive QA setting.
Our method extracts the sentence containing the ground truth answer span, i.e., the minimal context; the performance of our method on the 6B \& 7B test sets significantly improved.
We recognize that it is hard to prove the generalization of our method because the test sets for the BioASQ dataset are too small and the variance of performance is relatively high.
However, we demonstrate our superior performance by reducing the task discrepancy of factoid type questions in 6B \& and 7B.
Although, we have achieved better performance of list type questions, reducing the discrepancy of context length distribution does not significantly affect.
We believe that given the objective function of list type questions, it needs further analyses to demonstrate the generalization of sequential transfer learning with fine-tuning NLI dataset.

\begin{table}[t]
\begin{adjustbox}{max width=\textwidth}
\begin{tabular}{clclclcc}
\toprule
\multirow{2}{*}{\# of Batches~} & \multicolumn{2}{c}{Yes/No} & \multicolumn{2}{c}{Factoid} & \multicolumn{2}{c}{List} & \multirow{2}{*}{~Macro Avg.} \\ \cmidrule{2-7}
 & \multicolumn{1}{c}{System Name} & \multicolumn{1}{c}{Macro F1} & \multicolumn{1}{c}{System Name} & \multicolumn{1}{c}{MRR} & \multicolumn{1}{c}{System Name} & \multicolumn{1}{c}{F1} &  \\ \midrule
\multirow{3}{*}{8B batch 1} &  Ours & \textbf{0.8663} &  Ours & \textbf{0.4438} &  Ours & \textbf{0.3718} & \textbf{0.5606} \\
 & FudanLabZhu1 & 0.4518 & FudanLabZhu1 & 0.4557 & FudanLabZhu1 & 0.3408 & 0.4161 \\ 
 & Umass\_czi\_4 & 0.5989 & Umass\_czi\_4 & 0.3005 & Umass\_czi\_4 & 0.3448 & 0.4147 \\ \midrule
\multirow{3}{*}{8B batch 2} & Ours & \textbf{0.8928} & Ours & \textbf{0.3533} &  Ours & \textbf{0.3798} & \textbf{0.5420} \\ 
 & UoT\_multitask\_learn & 0.7000 & UoT\_multitask\_learn & 0.2800 & UoT\_multitask\_learn & 0.4108 & 0.4636 \\ 
 & FudanLabZhu4 & 0.6303 & FudanLabZhu4 & 0.2900 & FudanLabZhu4 & 0.4678 & 0.4627 \\ \midrule
\multirow{3}{*}{8B batch 3} & Umass\_czi\_4 & 0.9016 & Umass\_czi\_4 & 0.3810 & Umass\_czi\_4 & 0.4522 & 0.5782 \\ 
 & Ours & \textbf{0.9028} & Ours & \textbf{0.3601} & Ours & \textbf{0.4520} & \textbf{0.5716} \\
 & pa-base & 0.8995 & pa-base & 0.3137 & pa-base & 0.4585 & 0.5572 \\ \midrule
\multirow{3}{*}{8B batch 4} & Ours & \textbf{0.7636} & Ours & \textbf{0.6078} & Ours & \textbf{0.4037} & \textbf{0.5917} \\ 
 & 91-initial-Bio & 0.7204 & 91-initial-Bio & 0.5735 & 91-initial-Bio & 0.3905 & 0.5615 \\ 
 & Features Fusion & 0.7097 & Features Fusion & 0.5745 & Features Fusion & 0.3625 & 0.5489 \\ \midrule
\multirow{3}{*}{8B batch 5} & Ours & \textbf{0.8518} & Ours & \textbf{0.5677} & Ours & \textbf{0.5582} & \textbf{0.6592} \\
& Parameters retrained & 0.7509 & Parameters retrained & 0.5938 & Parameters retrained & 0.4004 & 0.5817 \\ 
& Features Fusion & 0.7509 & Features Fusion & 0.6115 & Features Fusion & 0.3810 & 0.5811 \\
\bottomrule
\end{tabular}
\end{adjustbox}
\vspace{2mm}
\caption{BioASQ 8B results obtained by the top three systems. The best scores were obtained from the BioASQ leaderboard (http://participants-area.bioasq.org/results/8b/phaseB/). We considered a system with different names as one system with the highest scores. We report the macro average scores obtained on all types of questions in the BioASQ dataset. Our systems are in \textbf{bold}.}
\vspace{-5mm}
\label{table:bioasq-8b-result}
\end{table}

\section{Analysis}
\subsubsection{Order of Sequential Transfer Learning}
\begin{table}[t]
\centering
\begin{adjustbox}{max width=\textwidth}
\begin{tabular}{clccc|ccc}
\toprule
\multicolumn{8}{c}{\multirow{2}{*}{Order Importance}} \\
\multicolumn{8}{c}{} \\ \midrule
\multicolumn{1}{c}{\multirow{2}{*}{~\# of Tasks~}} & \multicolumn{1}{c}{\multirow{2}{*}{~Sequence of Transfer Learning~~}} & \multicolumn{3}{c}{Factoid (\%)} & \multicolumn{3}{|c}{List (\%)} \\ \cmidrule{3-8} 
\multicolumn{1}{c}{} & \multicolumn{1}{c}{} & ~SAcc & ~LAcc & ~MRR & ~Prec & ~Recall & ~F1~ \\ \midrule
\multirow{3}{*}{6B Test} & BioBERT-SQuAD-BioASQ & 39.80 & 57.82 & 47.22 & 45.02 & \textbf{47.69} & 42.34 \\ 
 & BioBERT-SQuAD-MNLI-BioASQ & \textbf{41.15} & 57.95 & 47.29 & 46.18 & 44.56 & 40.98 \\
 & BioBERT-MNLI-SQuAD-BioASQ & 38.80 & \textbf{61.34} & \textbf{47.42} & \textbf{46.60} & 47.01 & \textbf{42.44} \\
 \midrule
\multirow{3}{*}{7B Test} & BioBERT-SQuAD-BioASQ & 41.95 & 58.30 & 48.66 & 61.32 & 52.83 & 52.36 \\ 
& BioBERT-SQuAD-MNLI-BioASQ & \textbf{43.31} & 58.69 & 49.24 & 60.77 & 50.74 & 50.72 \\ 
& BioBERT-MNLI-SQuAD-BioASQ & 42.22 & \textbf{61.06} & \textbf{49.85} & \textbf{61.46} & \textbf{54.62} & \textbf{54.19} \\ \bottomrule
\end{tabular}
\end{adjustbox}
\vspace{2mm}
\caption{Experiments on the importance of the order of sequential transfer learning. The metrics used for measuring performance on factoid-type questions are strict accuracy (SAcc), lenient accuracy (LAcc), and mean reciprocal rank (MRR). The metrics used for evaluating performance on list-type questions are precision (Prec), recall (Recall), and macro F1 (F1). The best score obtained in each task is in \textbf{bold}.}
\label{table:order experiment}
\end{table}

The BioASQ Challenge Task 8B (Phase B) results are shown in Table~\ref{table:bioasq-8b-result}. Each team was allowed to submit up to five systems with different combinations of features. 
The 8B ground truth answers are not available so we could not use them for manually evaluating our proposed method. Thus, we report the scores from the leaderboard.\footnote{http://participants-area.bioasq.org/results/8b/phaseB/}

In this ablation study, we explore the importance of the order of sequential transfer learning. The results are shown in Table~\ref{table:order experiment}.
We found that fine-tuning BioBERT on the MNLI dataset improved its performance on factoid type questions.
On the other hand, its performance on list type questions improved when the objective function of fine-tuned tasks was similar to that of the BioASQ task. 
In other words, BioBERT needs to be fine-tuned on the SQuAD dataset after fine-tuning it on the MNLI dataset.

\subsubsection{Unanswerable rate of the Extractive QA Setting}

\begin{table}
\centering
\begin{adjustbox}{max width=\textwidth}
\begin{tabular}{llcccccc}
\toprule
\multicolumn{2}{c}{Type~} & 7B Batch1~ & 7B Batch2~ & 7B Batch3~ & 7B Batch4~ & 7B Batch5~ & 7B Total \\ \midrule
\multicolumn{2}{l}{Factoid} & 0.359 (14/39) & 0.120 (3/25) & 0.310 (9/29) & 0.118 (4/34) & 0.229 (8/35) & 0.216 (35/162) \\ \midrule
\multicolumn{2}{l}{List} & 0.083 (1/12) & 0.235 (4/17) & 0.200 (5/25) & 0.136 (3/22) & 0.500 (6/12) & 0.204 (18/88) \\ \bottomrule
\end{tabular}
\end{adjustbox}
\vspace{2mm}
\caption{Statistics of the unanswerable rate in the extractive QA setting. The cases where \textit{Ground Truth Answer does not exactly match the context of the Human Annotated Corpus (Snippet)}. The unanswerable rate is related to the upper-bound performance of our proposed method.}
\vspace{-5mm}
\label{tab:human annotation}
\end{table}

So far, the experiments were performed in the extractive QA setting.
We manually analyzed differences between the answer span and the context of the human annotated corpus from the BioASQ Challenge Task 7B (Phase B) test set.
We used the test set instead of the training set for measuring the unanswerable rate of the extractive QA setting for the following two reasons.
First, we wanted to measure the upper-bound performance of our proposed method.
Second, the training and test data of the BioASQ dataset are similar to those of the dataset from the previous year.
Table~\ref{tab:human annotation} shows the unanswerable rate of all batch results of the 7B test set which contains only factoid and list type questions.
We calculated the unanswerable rate of the extractive QA setting using the rule \textit{Ground Truth Answer does not exactly match the context of the Human Annotated Corpus (Snippet)}. 
The rule applies to the following cases: no exact match, lowercase match, additional phrase added, and different type of blank space between the exact answer and snippet.
In Table~\ref{table:limit-human-annotation}, we randomly sample such cases.
Due to the lack of space, we provide more examples of cases at our url~\footnote{https://github.com/dmis-lab/bioasq8b/tree/master/human-eval}.
In here, we use the extractive QA setting to measure the upper-bound performance of our method. 
We hope our analysis is helpful in designing experimental settings.


\begin{table}
\centering
\begin{adjustbox}{max width=\textwidth}
\begin{tabular}{cl}
\toprule
\multicolumn{2}{c}{\multirow{2}{*}{Limitations of the Supervised Setting}} \\ \multicolumn{2}{c}{}  \\ \midrule
\multicolumn{1}{c}{Type} & \multicolumn{1}{c}{ID - Question - Context - Answer} \\ \midrule
Factoid  & \begin{tabular}[c]{@{}l@{}}
ID: 5c531d8f7e3cb0e231000017 \\
Question: What causes Bathing suit Ichthyosis(BSI)? \\ 
Ground Truth Answer: \textbf{transglutaminase-1 gene (TGM1) mutations} \\ 
Context: Bathing suit ichthyosis (BSI) is an uncommon phenotype classified as a minor \\
variant of autosomal recessive congenital ichthyosis (ARCI). OBJECTIVES: We report a case of \\
BSI in a 3-year-old Tunisian girl with a novel \textbf{mutation of the transglutaminase 1 gene (TGM1)} \\
\end{tabular} \\ \midrule
List & \begin{tabular}[c]{@{}l@{}}
ID: 5c5214207e3cb0e231000003 \\
Question: List potential reasons regarding why potentially important genes are ignored \\
Ground Truth Answer: \textbf{Identifiable chemical properties, Identifiable physical properties}, \\ \textbf{Identifiable} \textbf{biological properties}, \underline{Knowledge about homologous genes from model organisms} \\ 
Context: Here, we demonstrate that these differences in attention can be explained, to a large \\
extent, exclusively from a small set of \textbf{identifiable chemical, physical, and biological properties} \\
of genes. Together with \underline{knowledge about homologous genes from model organisms}, these \\ 
features allow us to accurately predict the number of publications on individual human \\
genes, the year of their first report, the levels of funding awarded by the National Institutes \\
of Health (NIH), and the development of drugs against disease-associated genes.
\end{tabular} \\ \bottomrule
\end{tabular}
\end{adjustbox}
\vspace{2mm}
\caption{Unanswerable questions of the extractive QA samples used for the BioASQ dataset. We used factoid- and list-type questions from the 7B test set. Context refers to a snippet in the human annotated corpus provided by the organizer of the BioASQ Challenge. No exact matches are in \textbf{bold} and exact matches in lowercase are \underline{underlined}.}
\vspace{-5mm}
\label{table:limit-human-annotation}
\end{table}

\section{Conclusion}
In this work, we used natural language inference (NLI) as a first step in fine-tuning BioBERT for biomedical question answering (QA). 
Training BioBERT to classify relationships between sentence pairs improved its performance in biomedical QA.
We empirically demonstrated that fine-tuning BioBERT on the NLI dataset improved its performance on the BioASQ dataset from the BioASQ Challenge.
We unified the distributions of context length to mitigate the discrepancy between NLI and biomedical QA. 
Furthermore, the order of sequential transfer learning is important when fine-tuning BioBERT.
Finally, when converting the format of the BioASQ dataset to the SQuAD format, we measured the unanswerable rate of the extractive QA setting where an answer does not exactly match the human annotated corpus.


\section*{Acknowledgements}
This research was supported by the National Research Foundation of Korea (NRF-2020R1A2C3010638, NRF-2016M3A9A7916996) and Korea Health Technology R\&D Project through the Korea Health Industry Development Institute (KHIDI), funded by the Ministry of Health \& Welfare, Republic of Korea (grant number: HR20C0021).

\bibliographystyle{splncs04}
\bibliography{bibgraphy}

\begin{thebibliography}{10}
\providecommand{\url}[1]{\texttt{#1}}
\providecommand{\urlprefix}{URL }
\providecommand{\doi}[1]{https://doi.org/#1}

\bibitem{williams2018broad}
Williams~et al., A.: A broad-coverage challenge corpus for sentence
  understanding through inference. In: Proceedings of the 2018 Conference of
  the NAACL: Human Language Technologies, Volume 1 (Long Papers) (2018)

\bibitem{wiese2017neural}
Wiese~et al., G.: Neural domain adaptation for biomedical question answering.
  In: Proceedings of the 21st Conference on CoNLL (2017)

\bibitem{levesque2012winograd}
Levesque~et al., H.: The winograd schema challenge. In: Thirteenth
  International Conference on the Principles of Knowledge Representation and
  Reasoning (2012)

\bibitem{phang2018sentence}
Phang~et al., J.: Sentence encoders on stilts: Supplementary training on
  intermediate labeled-data tasks. arXiv preprint arXiv:1811.01088  (2018)

\bibitem{oita2019semantically}
Oita~et al., M.: Semantically corroborating neural attention for biomedical
  question answering. In: ECML PKDD (2019)

\bibitem{telukuntla2019uncc}
Telukuntla~et al., S.K.: Uncc biomedical semantic question answering systems.
  bioasq: Task-7b, phase-b. In: ECML PKDD (2019)

\bibitem{hosein2019measuring}
Hosein~et al., S.: Measuring domain portability and errorpropagation in
  biomedical qa. arXiv preprint arXiv:1909.09704  (2019)

\bibitem{alsentzer2019publicly}
Alsentzer, E., Murphy, J., Boag, W., Weng, W.H., Jindi, D., Naumann, T.,
  McDermott, M.: Publicly available clinical bert embeddings. In: Proceedings
  of the 2nd Clinical Natural Language Processing Workshop (2019)

\bibitem{beltagy2019scibert}
Beltagy, I., Lo, K., Cohan, A.: Scibert: A pretrained language model for
  scientific text. In: Proceedings of the 2019 Conference on EMNLP-IJCNLP

\bibitem{bowman2015large}
Bowman, S., Angeli, G., Potts, C., Manning, C.D.: A large annotated corpus for
  learning natural language inference. In: Proceedings of the 2015 Conference
  on EMNLP

\bibitem{chen2020recall}
Chen, S., Hou, Y., Cui, Y., Che, W., Liu, T., Yu, X.: Recall and learn:
  Fine-tuning deep pretrained language models with less forgetting. arXiv
  preprint arXiv:2004.12651  (2020)

\bibitem{clark2019boolq}
Clark, C., Lee, K., Chang, M.W., Kwiatkowski, T., Collins, M., Toutanova, K.:
  Boolq: Exploring the surprising difficulty of natural yes/no questions. In:
  Proceedings of the 2019 Conference of the NAACL: Human Language Technologies,
  Volume 1 (Long and Short Papers) (2019)

\bibitem{devlin2019bert}
Devlin, J., Chang, M.W., Lee, K., Toutanova, K.: Bert: Pre-training of deep
  bidirectional transformers for language understanding. In: Proceedings of the
  2019 Conference of the NAACL: Human Language Technologies (2019)

\bibitem{dimitriadis2019yes}
Dimitriadis, D., Tsoumakas, G.: Yes/no question answering in bioasq 2019. In:
  ECML PKDD (2019)

\bibitem{howard2018universal}
Howard, J., Ruder, S.: Universal language model fine-tuning for text
  classification. In: Proceedings of the 56th Annual Meeting of the ACL (Volume
  1: Long Papers) (2018)

\bibitem{jin2019probing}
Jin, Q., Dhingra, B., Cohen, W., Lu, X.: Probing biomedical embeddings from
  language models. In: Proceedings of the 3rd Workshop on Evaluating Vector
  Space Representations for NLP (2019)

\bibitem{kim2019neural}
Kim, D., Lee, J., So, C.H., Jeon, H., Jeong, M., Choi, Y., Yoon, W., Sung, M.,
  Kang, J.: A neural named entity recognition and multi-type normalization tool
  for biomedical text mining. IEEE Access  (2019)

\bibitem{kim2019probing}
Kim, N., Patel, R., Poliak, A., Xia, P., Wang, A., McCoy, T., Tenney, I., Ross,
  A., Linzen, T., Van~Durme, B., et~al.: Probing what different nlp tasks teach
  machines about function word comprehension. In: Proceedings of the Eighth
  Joint Conference on Lexical and Computational Semantics (* SEM 2019) (2019)

\bibitem{lan2019albert}
Lan, Z., Chen, M., Goodman, S., Gimpel, K., Sharma, P., Soricut, R.: Albert: A
  lite bert for self-supervised learning of language representations. arXiv
  preprint arXiv:1909.11942  (2019)

\bibitem{lee2019biobert}
Lee, J., Yoon, W., Kim, S., Kim, D., So, C., Kang, J.: Biobert: a pre-trained
  biomedical language representation model for biomedical text mining.
  Bioinformatics (Oxford, England)  (2019)

\bibitem{liu-etal-2019-linguistic}
Liu, N.F., Gardner, M., Belinkov, Y., Peters, M.E., Smith, N.A.: Linguistic
  knowledge and transferability of contextual representations. In: Proceedings
  of the 2019 Conference of the NAACL: Human Language Technologies, Volume 1
  (Long and Short Papers) (2019)

\bibitem{long2015learning}
Long, M., Cao, Y., Wang, J., Jordan, M.I.: Learning transferable features with
  deep adaptation networks. arXiv preprint arXiv:1502.02791  (2015)

\bibitem{min2018efficient}
Min, S., Zhong, V., Socher, R., Xiong, C.: Efficient and robust question
  answering from minimal context over documents. In: Proceedings of the 56th
  Annual Meeting of the ACL (Volume 1: Long Papers) (2018)

\bibitem{mou2016transferable}
Mou, L., Meng, Z., Yan, R., Li, G., Xu, Y., Zhang, L., Jin, Z.: How
  transferable are neural networks in nlp applications? In: Proceedings of the
  2016 Conference on EMNLP

\bibitem{peng2019transfer}
Peng, Y., Yan, S., Lu, Z.: Transfer learning in biomedical natural language
  processing: An evaluation of bert and elmo on ten benchmarking datasets. In:
  Proceedings of the 18th BioNLP Workshop and Shared Task (2019)

\bibitem{peters2018deep}
Peters, M., Neumann, M., Iyyer, M., Gardner, M., Clark, C., Lee, K.,
  Zettlemoyer, L.: Deep contextualized word representations. In: Proceedings of
  the 2018 Conference of the NAACL: Human Language Technologies, Volume 1 (Long
  Papers) (2018)

\bibitem{raffel2019exploring}
Raffel, C., Shazeer, N., Roberts, A., Lee, K., Narang, S., Matena, M., Zhou,
  Y., Li, W., Liu, P.J.: Exploring the limits of transfer learning with a
  unified text-to-text transformer. arXiv preprint arXiv:1910.10683  (2019)

\bibitem{rajpurkar2016squad}
Rajpurkar, P., Zhang, J., Lopyrev, K., Liang, P.: Squad: 100,000+ questions for
  machine comprehension of text. In: Proceedings of the 2016 Conference on
  EMNLP

\bibitem{resta2019transformer}
Resta, M., Arioli, D., Fagnani, A., Attardi, G.: Transformer models for
  question answering at bioasq 2019. In: ECML PKDD (2019)

\bibitem{romanov2018lessons}
Romanov, A., Shivade, C.: Lessons from natural language inference in the
  clinical domain. In: Proceedings of the 2018 Conference on EMNLP

\bibitem{ruder2019neural}
Ruder, S.: Neural transfer learning for natural language processing. Ph.D.
  thesis (2019)

\bibitem{talmor2019multiqa}
Talmor, A., Berant, J.: Multiqa: An empirical investigation of generalization
  and transfer in reading comprehension. In: Proceedings of the 57th Annual
  Meeting of the ACL (2019)

\bibitem{tsatsaronis2015overview}
Tsatsaronis, G., Balikas, G., Malakasiotis, P., Partalas, I., Zschunke, M.,
  Alvers, M.R., Weissenborn, D., Krithara, A., Petridis, S., Polychronopoulos,
  D., et~al.: An overview of the bioasq large-scale biomedical semantic
  indexing and question answering competition. BMC bioinformatics  (2015)

\bibitem{vu2020exploring}
Vu, T., Wang, T., Munkhdalai, T., Sordoni, A., Trischler, A., Mattarella-Micke,
  A., Maji, S., Iyyer, M.: Exploring and predicting transferability across nlp
  tasks

\bibitem{wang2018glue}
Wang, A., Singh, A., Michael, J., Hill, F., Levy, O., Bowman, S.: Glue: A
  multi-task benchmark and analysis platform for natural language
  understanding. In: Proceedings of the 2018 EMNLP Workshop BlackboxNLP:
  Analyzing and Interpreting Neural Networks for NLP (2018)

\bibitem{yoon2019pre}
Yoon, W., Lee, J., Kim, D., Jeong, M., Kang, J.: Pre-trained language model for
  biomedical question answering. arXiv preprint arXiv:1909.08229  (2019)

\bibitem{yosinski2014transferable}
Yosinski, J., Clune, J., Bengio, Y., Lipson, H.: How transferable are features
  in deep neural networks? In: Advances in NIPS (2014)

\end{thebibliography}

\appendix
\section{Dataset Statistic}
\begin{table}
\centering
\begin{tabular}{clcccccc}
\toprule
\multicolumn{2}{c}{\textbf{MNLI}} & \multicolumn{3}{c}{Train} &  \multicolumn{2}{c}{Dev} \\ 
\multicolumn{2}{c}{Original} & \multicolumn{3}{c}{392,702} & \multicolumn{2}{c}{9,815} \\ \midrule
\multicolumn{2}{c}{\textbf{SQuAD v1.1}} & \multicolumn{3}{c}{Train} &  \multicolumn{2}{c}{Dev} \\
\multicolumn{2}{c}{Original} & \multicolumn{3}{c}{87,412} & \multicolumn{2}{c}{10,570} \\ 
\multicolumn{2}{c}{Snippet} & \multicolumn{3}{c}{82,280} & \multicolumn{2}{c}{9,986} \\ \midrule
\multicolumn{2}{c}{\textbf{SQuAD v2.0}} & \multicolumn{3}{c}{Train} &  \multicolumn{2}{c}{Dev} \\ 
\multicolumn{2}{c}{Original} & \multicolumn{3}{c}{130,319} & \multicolumn{2}{c}{11,873} \\ \midrule
\multicolumn{2}{c}{\textbf{BioASQ}} & \multicolumn{2}{c}{6B} & \multicolumn{2}{c}{7B} & \multicolumn{2}{c}{8B} \\  
Type & \multicolumn{1}{c}{Data Strategy~} & Train & Test~ & Train & Test~ & Train & Test~ \\ \midrule
Yes/No & Snippet-as-is & 9,421 & 127 & 10,560 & 140 & 11,531 & 152\\ \midrule
\multirow{3}{*}{Factoid} & Full-Abstract & 7,911 & & 9,403 & & 10,147 & \\ 
& Appended-Snippet & 5,953 & 161 & 7,179 & 162 & 7,896 & 151 \\ 
& Snippet-as-is & 3,512 & & 4,231 & & 4,759 &  \\ \midrule
\multirow{3}{*}{List} & Full-Abstract & 14,008 & & 15,719 & & 16,879 & \\ 
& Appended-Snippet & 10,878 & 81 & 12,184 & 88 & 13,251 & 75\\ 
& Snippet-as-is & 6,922 & & 7,865 & & 8,676 & \\ \bottomrule
\end{tabular}
\vspace{2mm}
\caption{Statistics of transferred dataset (MNLI \& SQuAD) and target dataset (BioASQ).}
\vspace{-5mm}
\label{table:statistics}
\end{table}
\section{Training Details}
\label{appendix:training-details}

We use BioBERT as learning biomedical entity representation. We utilize a single NVIDIA Titan RTX (24GB) GPU to fine-tune the sequence of transfer learning. In MNLI task, we use hyperparameters suggested by Hugging Face.\footnote{https://github.com/huggingface/transformers/tree/master/examples/text-classification} For fine-tuning, we select the batch size as 12, 24 and a learning rate is within range 1e-6 to 9e-6. In post-processing, we use the abbreviation resolution module called Ab3P\footnote{https://github.com/ncbi-nlp/Ab3P} to remove the same answer appearance with a different form. 

\section{Various NLI datasets}
\begin{table}
\centering
\begin{tabular}{lcccc|ccc|ccc}
\toprule
\multicolumn{1}{c}{} & \multicolumn{4}{c|}{Yes/No (\%)} & \multicolumn{3}{c|}{Factoid (\%)} & \multicolumn{3}{c}{List (\%)} \\ 
\cmidrule{2-5} \cmidrule{6-8} \cmidrule{9-11}
\multicolumn{1}{c}{Model} & ~Accuracy & Yes F1 & No F1 & ~Macro F1~ & ~SAcc~ & LAcc & ~MRR~ & ~Prec~ & Recall & F1 \\ \midrule
SQuAD & 85.18 & 90.04 & 68.96 & 79.50 & 39.80 & 57.82 & 47.22 & 45.02 & 47.69 & 42.34 \\ \midrule
MNLI & \textbf{88.57} & 92.12 & \textbf{77.98} & \textbf{85.05} & 38.80 & \textbf{61.34} & \textbf{47.42} & \textbf{47.86} & 46.89 & \textbf{43.33} \\ \midrule
SNLI & 88.51 & \textbf{92.17} & 77.47 & 84.82 & 39.11 & 58.23 & 46.96 & 44.42 & \textbf{48.16} & 42.20 \\ \midrule
MedNLI & 77.81 & 85.24 & 52.32 & 68.78 & \textbf{40.05} & 57.66 & 47.14 & 45.56 & 47.31 & 42.72 \\ 
\bottomrule
\end{tabular}
\vspace{2mm}
\caption{Experiments of various NLI datasets evaluated on BioASQ 6B (Phase B). The experiments are considered as a first step of the sequential transfer learning. The model of Yes/No type are fine-tuned as same as Table~\ref{table:yes-no type experiment}. The model of Factoid and List type are fine-tuned as same as Table~\ref{table:task discrepancy}. The best score obtained in each task is in \textbf{bold}.}
\label{table:nli dataset experiments 6B}
\end{table}

\begin{table}
\centering
\begin{tabular}{lcccc|ccc|ccc}
\toprule
\multicolumn{1}{c}{} & \multicolumn{4}{c|}{Yes/No (\%)} & \multicolumn{3}{c|}{Factoid (\%)} & \multicolumn{3}{c}{List (\%)} \\ 
\cmidrule{2-5} \cmidrule{6-8} \cmidrule{9-11}
\multicolumn{1}{c}{Model} & ~Accuracy & Yes F1 & No F1 & ~Macro F1~ & ~SAcc~ & LAcc & ~MRR~ & ~Prec~ & Recall & F1 \\ \midrule
SQuAD & 85.95 & 89.90 & 73.44 & 81.67 & 41.95 & 58.30 & 48.66 & 61.32 & 52.83 & 52.36 \\ \midrule
MNLI & \textbf{89.45} & \textbf{92.75} & \textbf{75.88} & \textbf{84.32} & \textbf{42.22} & \textbf{61.06} & \textbf{49.85} & \textbf{61.46} & \textbf{54.62} & \textbf{54.19} \\ \midrule
SNLI & 85.40 & 90.11 & 66.95 & 78.53 & 41.84 & 60.03 & 49.31 & 56.20 & 48.07 & 47.70 \\ \midrule
MedNLI & 78.67 & 85.38 & 49.20 & 67.29 & 41.45 & 60.55 & 49.05 & 58.40 & 48.17 & 48.25 \\ 
\bottomrule
\end{tabular}
\vspace{2mm}
\caption{Experiments of various NLI datasets evaluated on BioASQ 7B (Phase B). The experiments are considered as a first step of the sequential transfer learning. The model of Yes/No type are fine-tuned as same as Table~\ref{table:yes-no type experiment}. The model of Factoid and List type are fine-tuned as same as Table~\ref{table:task discrepancy}. The best score obtained in each task is in \textbf{bold}.}
\label{table:nli dataset experiments 7B}
\end{table}

\end{document}